\documentclass[10pt,letterpaper]{article}

\usepackage[final]{cogsci}
\usepackage{subcaption}
\cogscifinalcopy 

\usepackage{graphicx}
\usepackage[
  style=apa,
  natbib=true,
  annotation=false,
]{biblatex}
\usepackage{float} 
\usepackage{booktabs}
\usepackage{multirow}
\usepackage{amsmath}
\usepackage{siunitx}

\usepackage{siunitx}

\usepackage{xcolor}

\newif\ifshowcomments
\showcommentstrue      

\usepackage{hyperref}

\definecolor{DLcolor}{RGB}{86, 180, 233}   
\definecolor{MScolor}{RGB}{0, 158, 115}    
\definecolor{EYcolor}{RGB}{230, 159, 0}    
\definecolor{ACcolor}{RGB}{204, 121, 167}  

\title{Human Adults and LLMs as Scientists: Who Benefits from Active Exploration?}

\author{Mandana Samiei $^{1, 2 *}$ \hspace{0.5em}Eunice Yiu $^{3 *}$ \hspace{0.5em}Anthony GX-Chen $^{4}$ \hspace{0.5em}Dongyan Lin $^{5}$ \hspace{0.5em}Jocelyn Shen $^{6}$ \\ \vspace{2mm}
\large\textbf{Blake A. Richards $^{1, 2, 7}$ \hspace{0.5em} Alison Gopnik $^{3}$ \hspace{0.5em} Doina Precup $^{1, 2}$} \vspace{2.3mm}

$^{1}$Mila - Quebec AI Institute \hspace{0.3em} 
$^{2}$McGill University \hspace{0.3em} 
$^{3}$University of California Berkeley \hspace{0.3em}
$^{4}$New York University \hspace{0.3em} \\ \vspace{2mm}
\large $^{5}$Meta FAIR \hspace{0.3em} 
\large $^{6}$MIT Media Lab \hspace{0.3em}
\large $^{7}$Montreal Neurological Institute \hspace{0.3em}
\\ \vspace{2mm}
$^*$Equal contribution. Correspondence: \texttt{mandana.samiei@mail.mcgill.ca, ey242@berkeley.edu} \vspace{-4mm}
} 

\begin{document}

\maketitle

\begingroup
\renewcommand\thefootnote{}
\footnotetext{\emph{Accepted at the 48th Annual Conference of the Cognitive Science Society (CogSci 2026)}.}
\endgroup

\vspace{-5mm}
\begin{abstract}
\vspace{-1mm}
A long-standing finding in the causal learning literature is that adults struggle to identify conjunctive causal rules, where an effect requires the simultaneous presence of multiple causes, while performing better in disjunctive settings. However, most demonstrations of this ``conjunctive handicap'' rely on passive observation paradigms with limited evidence, where learners have no control over evidence generation. This paper asks whether this bias persists when adults are granted agency through active exploration. Using a modified ``blicket detector'' task, adult participants freely intervened to identify causal objects under conjunctive or disjunctive rule structures. We show that active exploration substantially improves adults' conjunctive causal reasoning, although conjunctive rules still require more tests to infer than disjunctive rules. We further compare human performance to a range of large language models in the same setting. While some state-of-the-art models approach human-level performance on hypothesis inference accuracy, they often exhibit less efficient exploration strategies and similar conjunctive-disjunctive performance gaps.

\vspace{-1.2mm}
\end{abstract}
\vspace{-1mm}
\section{Introduction}

Understanding how agents infer causal structure is a central problem in cognitive science. Causal learning supports prediction, explanation, intervention, and scientific discovery \citep{woodward2003making}. Developmental work shows that even young children can infer latent causal variables, reason about unobserved mechanisms, and distinguish correlation from causation from limited evidence \citep{gopnik2000detecting,sobel2004children,schulz2007preschool}.

\noindent Much of this work has relied on the ``blicket detector'' paradigm~\citep{gopnik2000detecting}, in which learners observe a machine that activates when certain objects, or combinations of objects, are placed on it. A striking and perhaps counterintuitive finding from this literature is that children sometimes outperform adults in learning abstract causal structure. Adults often default to ``disjunctive'' (OR) rules, whereas preschool-aged children readily infer ``conjunctive'' (AND)  rules when the training evidence supports them \citep{lucas2014children}. This adult ``conjunctive handicap'' has been observed across cultures and socioeconomic contexts \citep{wente2019causal}.


\noindent However, these demonstrations of adult failure in conjunctive causal reasoning have exclusively relied on \emph{passive} learning paradigms, where learners observe a fixed sequence of evidence without control over which interventions are performed. This is important because causal learning is shaped by intervention and exploration. Children learn more effectively when they can design their own tests, and adults also learn worse when interventions are imposed rather than self-directed \citep{schulz2007preschool,sobel2006importance,sobel2013interventions}. 

\noindent These findings raise a key question: Are adults actually poor conjunctive causal learners, or do they struggle because passive evidence presentation prevents them from generating and evaluating the right evidence to update their priors? 

\begin{figure*}[t]
    \centering
    \includegraphics[width=\textwidth, trim=0cm 2.1cm 3cm 0cm, clip]{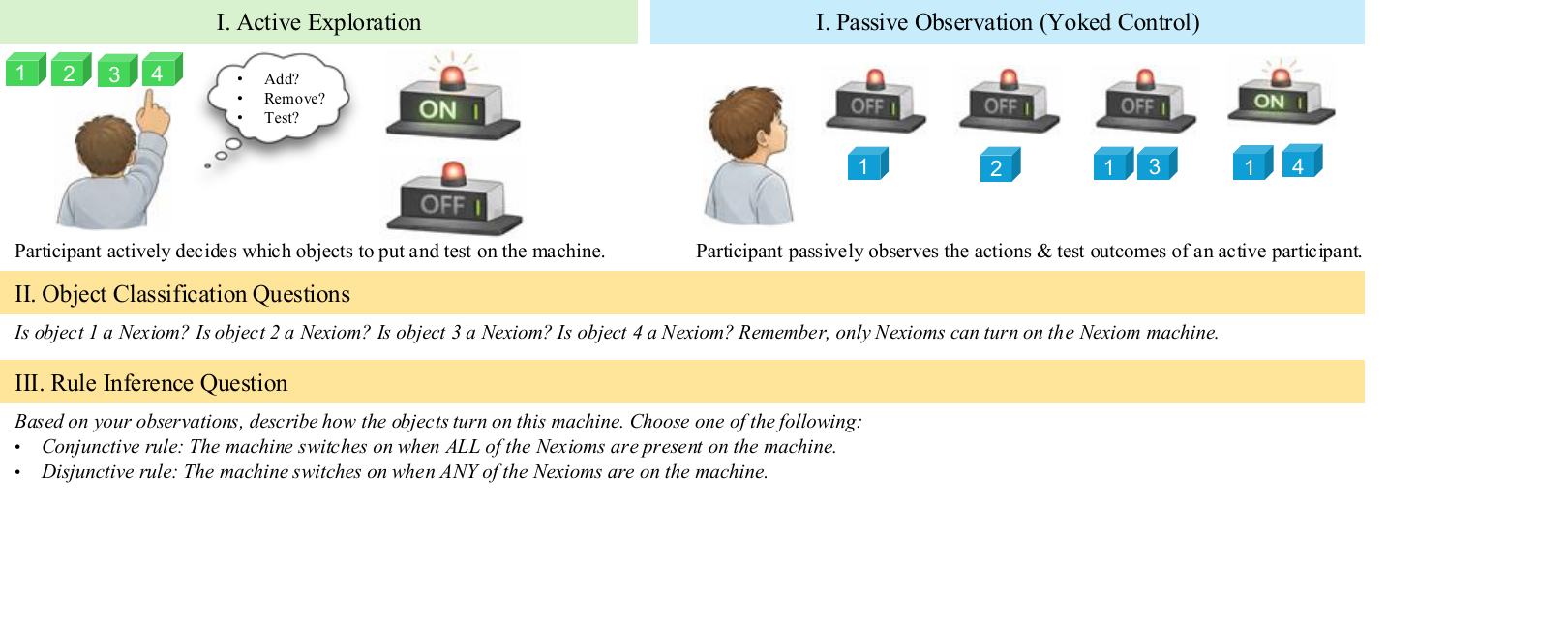}
    \vspace{-25pt}
    \caption{\textbf{Test structure in the Active Exploration vs. Passive Observation conditions.}
    In the active exploration condition, participants can click to add or remove four individual objects from the Nexiom machine, and explicitly test the current combination to observe whether the machine switches ON or OFF. In the passive observation condition, participants are each paired to an active exploration participant (yoked control). They do not perform any actions on their own, but rather observe the actions and test outcomes of the active exploration participants. Participants in both conditions are then asked to classify which object(s) are ``Nexioms'' and the rule under which the objects operate (conjunctive or disjunctive) to activate the machine.\vspace{-0.8em}
    }
    \label{fig:exptflow}
\end{figure*}

\noindent We address this question using a novel ``nexiom detector'' task, a blicket-style causal learning paradigm designed to minimize familiar task-specific priors.  Adult participants could test any object or combination of objects, and decide when they had enough evidence to infer both the causal objects and the underlying rule, allowing us to compare active and passive causal learning. 


\noindent We find that active exploration substantially changes adult performance. In contrast to prior passive studies, active adults show strong performance on conjunctive causal rules and generated relatively small but highly informative sets of tests. This suggests that adult failures in conjunctive causal reasoning \citep{lucas2014children} reflect constraints of passive evidence presentation, not a fixed limitation in causal competence.

\noindent We further showed that just generating informative tests was not enough. Participants who proposed interventions but observed outcomes from someone else’s tests performed as poorly as the passive learners. This suggests that active exploration helps most when intervention choices are tightly coupled to their own contingent outcomes.

\noindent Finally, we situate these findings in a broader computational context by comparing adults to large language models (LLMs) on the same causal discovery tasks. LLMs are increasingly evaluated as general-purpose reasoning agents capable of hypothesis generation and experimentation \citep{piriyakulkij2024doing,zhang2021acre}, and recent work has placed them in blicket-style environments where they select interventions to observe outcomes \citep{gx2025language}. This makes them a natural comparison case for examining whether active intervention is sufficient for successful causal discovery. We find that LLMs do not consistently gain from choosing their own interventions and still lag behind the top-performing human explorers on conjunctive rules. \\ 

\noindent Together, these results suggest that successful causal discovery depends critically on maintaining a tight coupling between self-generated interventions and their contingent outcomes. Active intervention alone is insufficient: effective causal learning requires adaptive search strategies and progressive hypothesis pruning, particularly in conjunctive environments.

\vspace{-3mm}
\section{Methods}

\paragraph{The Nexiom Detector Software.} We developed a custom web-based platform, Nexiom Text Adventure \footnote{\url{https://nexiom-text-game.streamlit.app/}}, to study active causal reasoning in human adults. The task is functionally equivalent to the classic blicket detector paradigm, but uses the novel term ``nexiom''  to minimize prior knowledge from the established blicket literature. The platform was implemented in Streamlit and supports structured experimental sessions involving text-based scenarios with interactive object selection and explicit causal testing, and records detailed behavioral data, including object selections, test sequences, test outcomes, response times, object identification accuracy, and rule inference judgments. Each session consists of a comprehension phase followed by a main test phase. \\

\begin{figure}[ht]
    \centering
    \includegraphics[width=\linewidth]{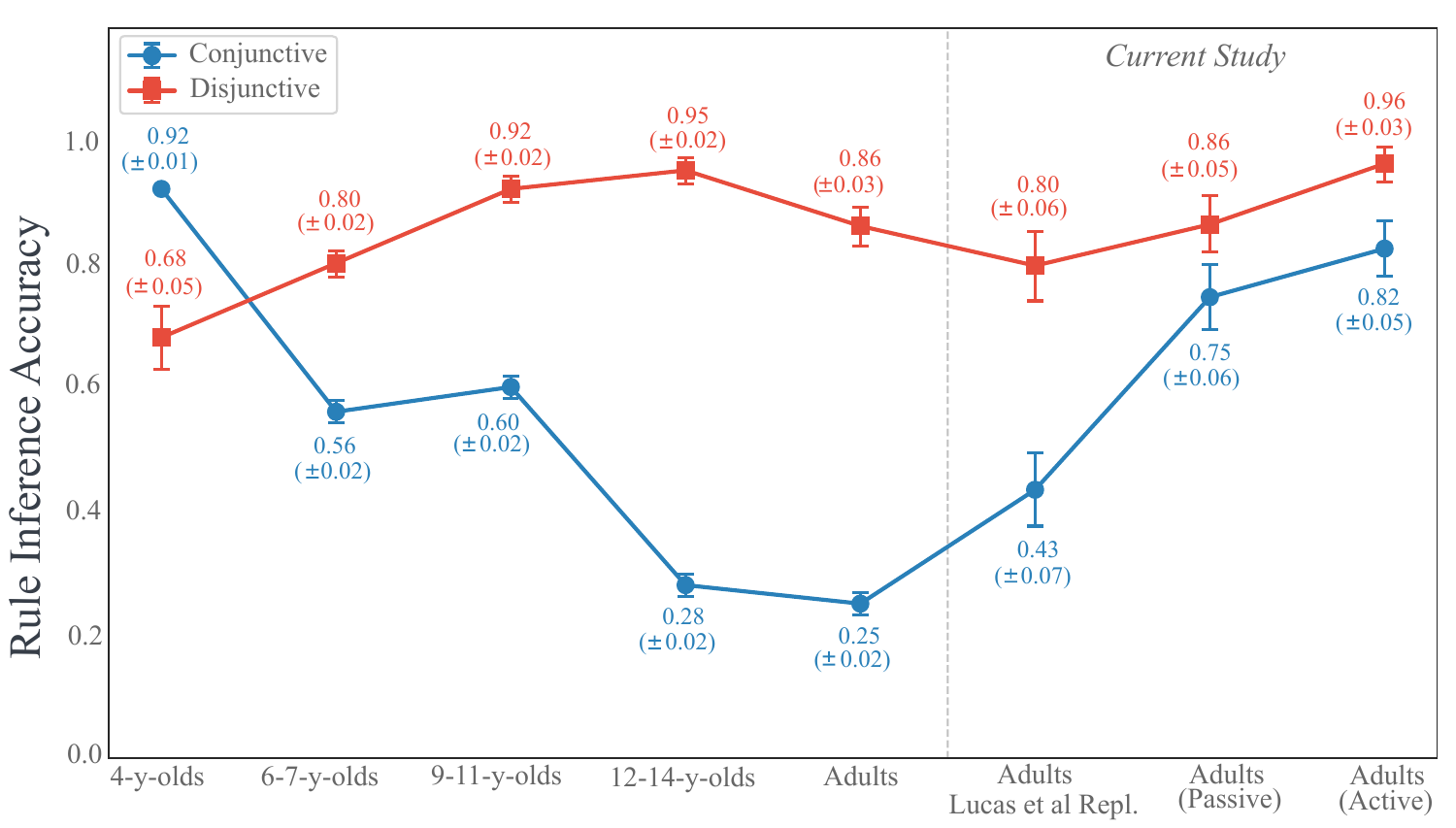}
    \vspace{-6.3mm}
    \caption{\textbf{Rule inference accuracy across age groups.} Performance of children and adults when given ambiguous passive-observation data in a prior study (\citet{gopnik2017changes} (left of dashed line) is compared with adults' performance in the current study (right of dashed line). \vspace{-1.2em}} 
    \label{fig:PNAS}
\end{figure}

\vspace{-3mm}
\noindent\textbf{Experiments.} Participants were assigned to either the "Active Exploration" or the  "Passive Observation" condition. Active Exploration participants engaged in unrestricted causal exploration to identify both the causal objects and the underlying causal rule. They could freely select objects, test arbitrary combinations, and decide when to stop exploring, whereas Passive Observation participants could only observe the actions and test outcomes of a paired Active Exploration participant (yoked control) to learn about the underlying causal hypothesis of the machine. After exploration, participants were first asked to identify which objects they believed were "nexioms" and could switch on the machine. Next, they were explicitly informed of two possible rule types (conjunctive vs.\ disjunctive), and were asked to select which rule they believed governed the machine. Figure~\ref{fig:exptflow} illustrates the experimental flow. We used the Passive Observation condition to replicate the ambiguous evidence paradigm used in prior work on conjunctive causal learning in adults \citep{lucas2014children,gopnik2017changes} (see Figure~\ref{fig:PNAS}). 
We additionally conducted a supplementary "Passive Proposer" experiment to isolate whether the benefits of active exploration arise from intervention planning itself or from receiving contingent feedback from one’s own interventions. Passive Proposers were allowed to actively generate hypotheses and propose interventions, but did not directly observe the outcomes of their own proposed tests. Instead, they received the outcomes generated by a matched Active Exploration participant.  Finally, we extended the same active exploration framework to several large language models. Models were allowed to sequentially select interventions by adding or removing objects, testing the resulting combination, and observe machine outcomes. In principle, we provide LLMs with access to the same action space as human learners. Model evaluations used the same conjunctive and disjunctive causal structures as the human experiments, enabling direct comparison of exploration behavior, object identification, and rule inference across agents.

\vspace{-3mm}

\paragraph{Data.} A total of 306 adult participants (age: 22--35 years, $M_{\mathrm{age}} = 30.41$, $SD_{\mathrm{age}} = 3.94$; 153 female, 153 male) were recruited on Prolific\footnote{https://www.prolific.com} to complete an online experiment in accordance with IRB-approved protocols. 102 participants were assigned to the Active Exploration condition, 102 were assigned to the Passive Observation condition, and another 102 were assigned to the Passive Proposer condition. Participants in both passive conditions were randomly paired with a unique Active Exploration participant to create a passive yoked-control setup. Only participants who successfully completed the experiment were included in the analyses. In each condition, participants were randomly assigned to one of two between-subjects causal rules in the test phase: $N = 51$ were assigned to the conjunctive condition and another $N = 51$\footnote{Throughout the paper, "top human" refers to the highest-performing human participants within each condition, defined as participants achieving perfect full-hypothesis inference accuracy.} were assigned to the disjunctive condition.
To compare human causal exploration with artificial agents under matched interaction settings, we conducted experiments with language models using 24 trials per rule type (conjunctive / disjunctive), resulting in 48 total trials per model. The 24 trials consisted of 6 unique causal configurations, corresponding to all possible combinations of 2 blickets selected from 4 objects, with each configuration evaluated four times.  We tested six language models spanning both reasoning-oriented and non-reasoning agents: \texttt{gpt-5}, \texttt{gpt-5-mini}, \texttt{gemini-2.5-flash}, \texttt{deepseek-reasoner}, \texttt{o4-mini}, and \texttt{deepseek-chat}. Across all models and rule conditions, this resulted in a total of 288 LLM evaluation trials (using temperature = $0.0$).  

\begin{figure}[ht]
    \centering
    \includegraphics[width=0.9\linewidth]{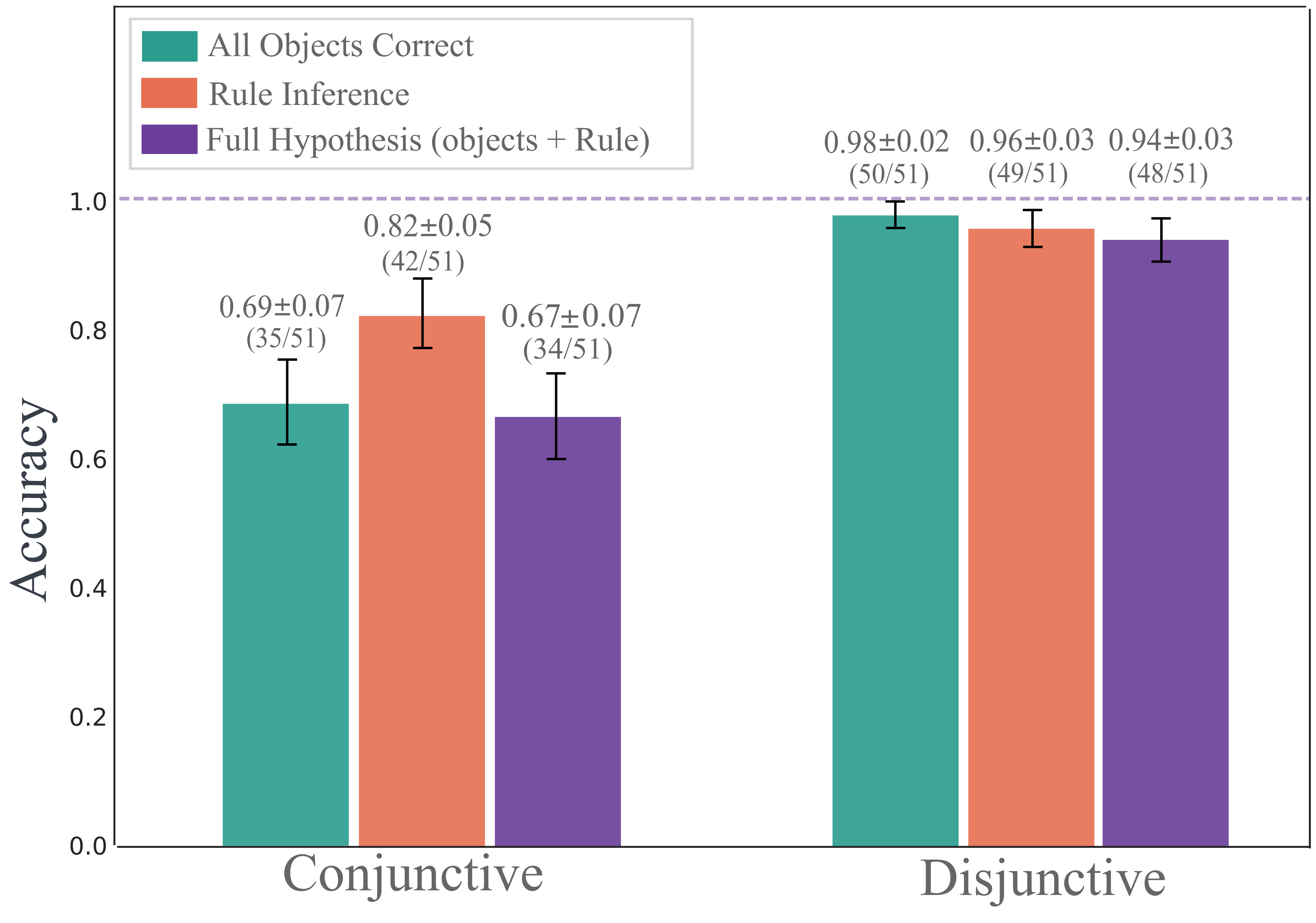}
    \vspace{-3mm}
    \caption{\textbf{Performance accuracy in conjunctive vs.\
    disjunctive causal inference with active exploration evidence.} Error bars are $\pm$ standard error of the mean across participants.\vspace{-5mm}}
    \label{fig:accuracies}
\end{figure}

\vspace{-3mm}
\paragraph{Evaluation Metrics.} We used three binary accuracy measures (scored 0 or 1) to evaluate participants’ causal understanding and excluded 4 additional participants with prior experience from all analyses. \textit{Object identification accuracy} is defined as the proportion of participants who selected the exact set of causal objects (``nexioms''). A trial was scored as correct only if the participant’s selected set matched the ground-truth nexioms exactly (e.g., if the correct nexioms were \texttt{[2,4]}, selecting any other combination was scored as incorrect). \textit{Rule inference accuracy} is defined as the proportion of participants who correctly identified the underlying causal rule (conjunctive vs.\ disjunctive). \textit{Full hypothesis accuracy} required both correct object identification and correct rule inference. For each metric and rule condition, we report mean accuracy and standard error (see Figure~\ref{fig:accuracies}). \\
We additionally report exploration-process measures derived from participants' test sequences. \textit{Cumulative information gain} quantities the reduction in uncertainty (in bits) accumulated across tests. Before each test, we compute the  number of hypotheses consistent with prior evidence; after observing the machine’s response (on/off), we recompute the number of remaining consistent hypotheses:
$\text{InfoGain}_t = \log_2(N_{t-1}) - \log_2(N_t) = \log_2\left(\frac{N_{t-1}}{N_t}\right)$, where $N_{t-1}$ is the number of hypotheses consistent before test $t$, and $N_t$ is the number of hypotheses consistent after test $t$.
Cumulative information gain up to test $k$ is defined as $\text{CumInfoGain}_k = \sum_{t=1}^{k} \text{InfoGain}_t$.  \textit{Number of hypotheses remaining} tracks how many full hypotheses are still consistent with the test outcomes after each test. The full hypothesis space comprises all pairs of (Nexiom set, rule).
We further compute the average number of tests conducted and the time spent per test across trials. These measures are reported separately for conjunctive and disjunctive conditions. The statistics presented in Tables~\ref{tab:exploration-all} and ~\ref{tab:exploration-success} do not consider the time spent in the comprehension phase. 
\vspace{-4mm}
\paragraph{Raw Data Availability.}
Behavioral data and analysis-ready summary files are available on \href{https://osf.io/fsvqb/overview?view_only=55a429715ee24f1bbccfb4d30bb7fe05}{the Open Science Framework (OSF)}. The repository includes raw interaction logs and processed summary data for the analyses here.

\vspace{-2mm}

\section{Results}

\begin{figure}[t]
    \centering
    \includegraphics[width=0.5\textwidth]{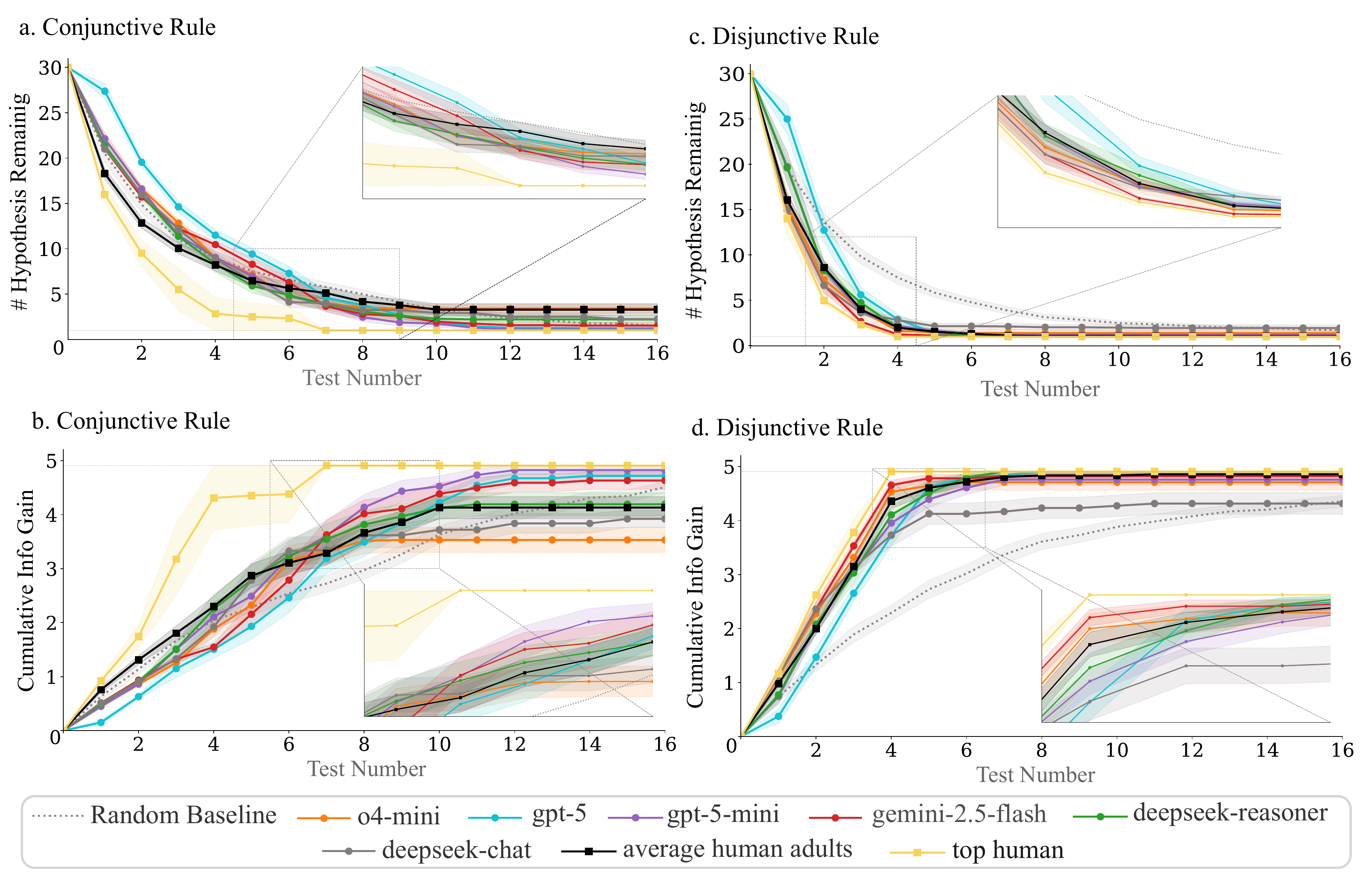}
    \vspace{-5mm}
    \caption{\textbf{Search efficiency.} Comparing hypothesis pruning \& information gain in conjunctive vs. disjunctive exploration.\vspace{-5mm}}
    \label{fig:hypothesis_infogain}
\end{figure}

\begin{figure}[ht]
    \centering
    \includegraphics[width=\linewidth]{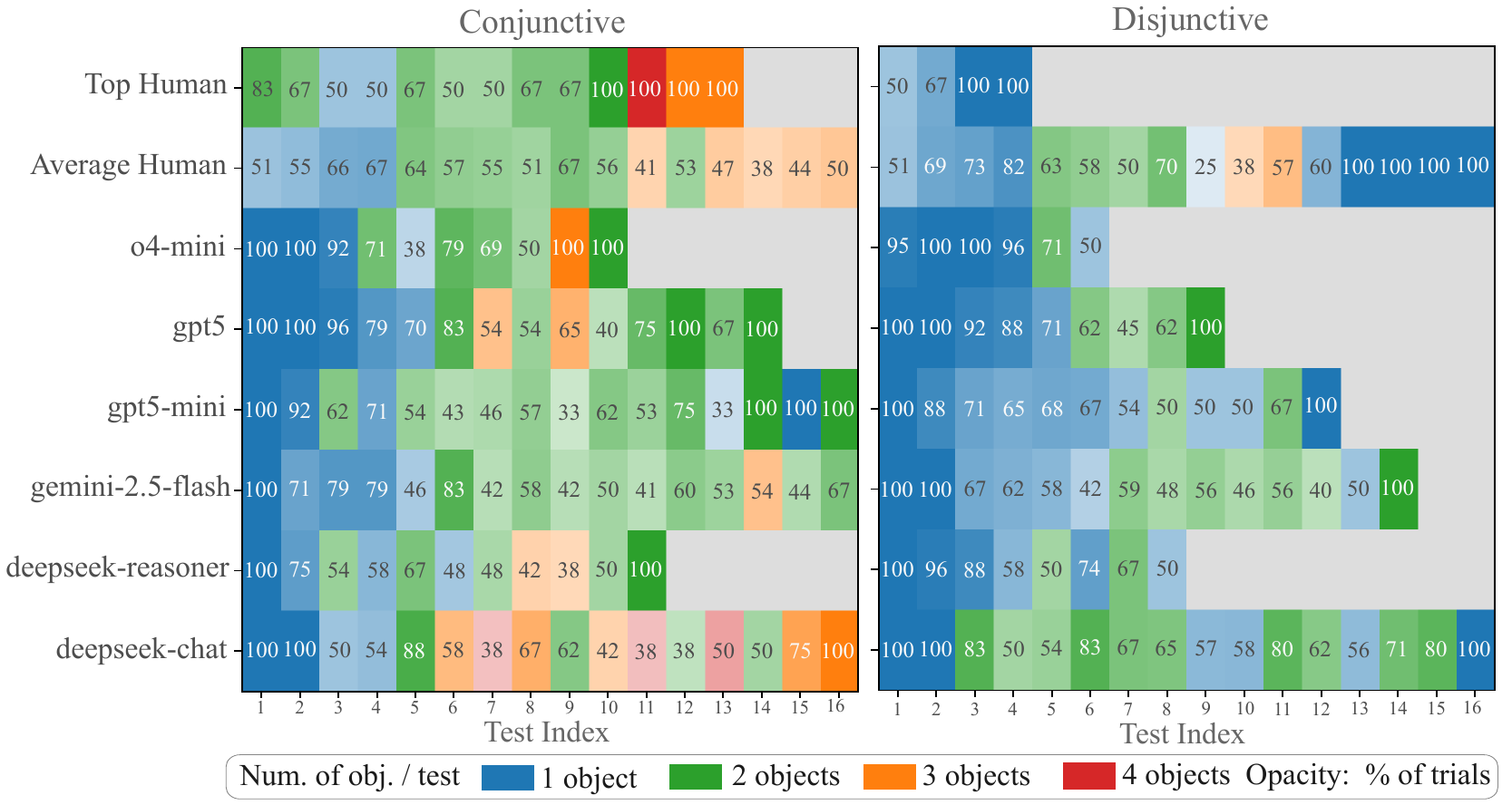}
    \vspace{-4mm}
    \caption{\textbf{Sequential search strategy across human and language agents during active exploration.} Each color represents the dominant number of objects per test index, with opacity indicating the percentage of trials that have that dominant number of objects within that index. Grey area shows the terminating point of search. \vspace{-5mm}}
    \label{fig:search_strategy}
\end{figure}

\begin{figure}[ht]
    \centering
\includegraphics[width=\linewidth]{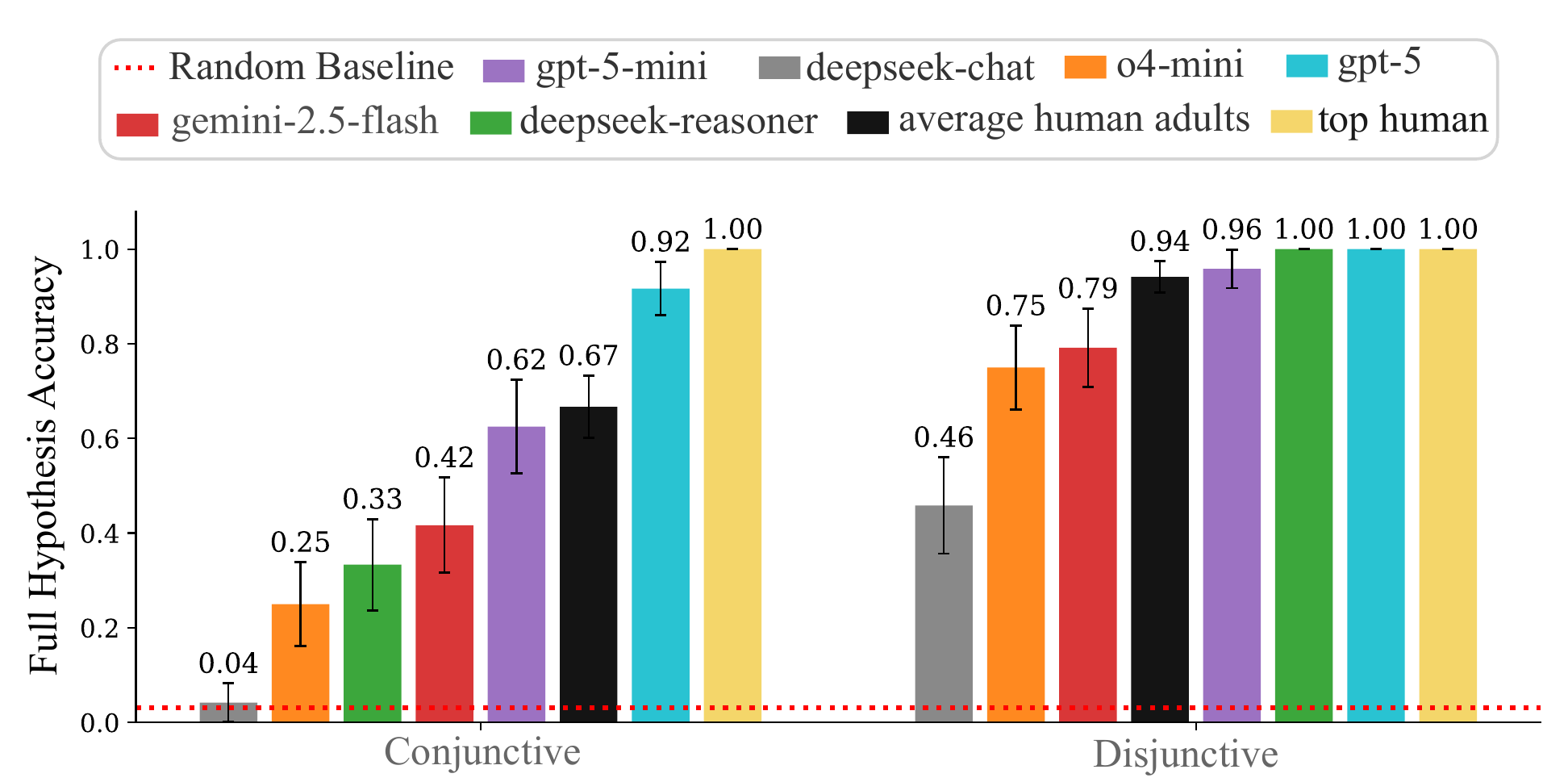}
    \vspace{-6mm}
    \caption{\textbf{Comparative analysis of causal reasoning across LLMs and Humans}. Error bars indicate ±SEM, and numeric labels above bars report mean accuracy.\vspace{-1mm}}
    \label{fig:llm-human-bar-active}
\end{figure}

\noindent Beyond overall accuracy, our results show that both human participants and language models do not interact with the Nexiom machine in a random or unguided manner. Instead, they engaged in systematic exploration designed to disambiguate competing causal solutions. The hypothesis space initially consists of 32 possible explanations.\footnote{For $N$ items and two rule types (conjunctive vs.\ disjunctive), the total hypothesis space is $2^{N+1} $.} Across both conjunctive and disjunctive conditions, participants and stronger models systematically reduced this space through targeted interventions, although conjunctive conditions required substantially longer exploration trajectories to achieve comparable reductions in uncertainty. As shown in Figures~\ref{fig:hypothesis_infogain}, ~\ref{fig:search_strategy}, and~\ref{fig:llm-human-bar-active}, successful causal discovery was associated not only with higher final accuracy, but also with more efficient hypothesis pruning, greater cumulative information gain, and more adaptive sequential search strategies over the course of exploration.
\vspace{-2mm}
\paragraph{Exploration Strategies, Passive vs.\ Active Learners.} Several recent developmental works \citep{pmlr-v177-kosoy22a, gopnik2017changes} asked whether learners can actively generate the data necessary to form correct hypotheses, rather than simply receiving that data from an experimenter. Our study addresses this by placing adults in an environment where they must determine the governing logic (disjunctive vs. conjunctive) through self-directed interventions. Figure~\ref{fig:hypothesis_infogain} shows that disjunctive environments permit rapid pruning of the hypothesis space, with both humans and stronger models eliminating most competing hypotheses within the first few tests. In contrast, conjunctive environments require substantially longer exploration trajectories to achieve comparable reductions in uncertainty. This asymmetry is also reflected in cumulative information gain: information accumulates rapidly in disjunctive settings, whereas conjunctive settings require more gradual evidence accumulation across a larger number of tests. The top-performing human adults nevertheless show the most efficient search behavior across all models, progressively narrowing the hypothesis space through targeted interventions. 
\vspace{-2mm}

\vspace{-2mm}
\begin{table}[ht]
\centering
\caption{\textbf{Comparison of exploration effort between humans and models} across conjunctive and disjunctive rules ($M \pm SE$).\vspace{-2mm}}
\label{tab:exploration-all}
\scriptsize 
\setlength{\tabcolsep}{2.5pt}
\begin{tabular}{l cc cc cc}
\toprule
& \multicolumn{2}{c}{\# Tests} & \multicolumn{2}{c}{\# Actions} & \multicolumn{2}{c}{Time/Test (s)} \\
\cmidrule(lr){2-3} \cmidrule(lr){4-5} \cmidrule(lr){6-7}
Agent & Conj. & Disj. & Conj. & Disj. & Conj. & Disj. \\
\midrule
Avg. human & 9.6 $\pm$ .6 & 6.4 $\pm$ .4 & 32.1 $\pm$ 2.4 & 20.8 $\pm$ 1.4 & 31.5 $\pm$ 5.2 & 23.5 $\pm$ 1.3 \\
Top human & 8.5 $\pm$ .99 & 3.8 $\pm$ .17 & 29.3 $\pm$ 5.8 & 11.2 $\pm$ 0.5 & 23.9 $\pm$ 5.3 & 24.6 $\pm$ 3.5 \\
\midrule
gpt-5 & 10.2 $\pm$ .4 & 7.7 $\pm$ .2 & 26.2 $\pm$ 1.1 & 21.0 $\pm$ .6 & 44.3 $\pm$ 4.6 & 43.0 $\pm$ 5.5 \\
gpt-5-mini & 10.8 $\pm$ .4 & 7.1 $\pm$ .5 & 27.5 $\pm$ 1.2 & 18.8 $\pm$ 1.1 & 30.1 $\pm$ 1.1 & 28.7 $\pm$ 1.0 \\
gemini-2.5-f & 13.5 $\pm$ .4 & 9.71 $\pm$ .4 & 36.5 $\pm$ 1.9 & 28.3 $\pm$ 1.4 & 9.6 $\pm$ .3 & 11.1 $\pm$ .3 \\
ds-reasoner & 8.2 $\pm$ .3 & 6.8 $\pm$ .2 & 22.0 $\pm$ 1.2 & 19.6 $\pm$ .6 & 65.2 $\pm$ 4.8 & 58.2 $\pm$ 4.8 \\
o4-mini & 6.7 $\pm$ .3 & 4.2 $\pm$ .2 & 24.2 $\pm$ 1.6 & 12.8 $\pm$ .5 & 24.4 $\pm$ 1.2 & 22.0 $\pm$ 1.5 \\
ds-chat & 12.3 $\pm$ .4 & 11.9 $\pm$ .6 & 38.7 $\pm$ 1.5 & 35.7 $\pm$ 1.8 & 5.5 $\pm$ .2 & 6.0 $\pm$ .2 \\

\bottomrule
\end{tabular}
\vspace{-1em}
\end{table}

\vspace{-1mm}

\vspace{-2mm}
\begin{table}[ht]
\centering
\caption{\textbf{Number of tests and time spent per test} across successful (both rule and objects) trials ($M \pm SE$).\vspace{-2mm}}
\label{tab:exploration-success}
\scriptsize 
\setlength{\tabcolsep}{2pt}
\begin{tabular}{l cc cc cc}
\toprule
& \multicolumn{2}{c}{$N$. Succ.} & \multicolumn{2}{c}{\# Tests} & \multicolumn{2}{c}{Time/test (s)} \\
\cmidrule(lr){2-3} \cmidrule(lr){4-5} \cmidrule(lr){6-7}
Agent & Conj. & Disj. & Conj. & Disj. & Conj. & Disj. \\
\midrule
Avg. human & 34 & 48 & 11.24 $\pm$ .53 & 6.48 $\pm$ .45 & 19.83 $\pm$ 1.30 & 23.27 $\pm$ 1.17 \\
Top human & 6 & 6 & 8.50 $\pm$ .99 & 3.83 $\pm$ .17 & 23.88 $\pm$ 5.26 & 24.62 $\pm$ 3.48 \\
\midrule
gpt-5 & 22 & 24 & 10.36 $\pm$ .4 & 7.67 $\pm$ .45 & 40.68 $\pm$ 2.92 & 43.05 $\pm$ 6.45 \\
gpt-5-mini & 15 & 23 & 10.87 $\pm$ .47 & 7.26 $\pm$ .45 & 28.07 $\pm$ 1.06 & 28.27 $\pm$ .98 \\
gemini-2.5-f & 10 & 19 & 13.8 $\pm$ .53 & 10.0 $\pm$ .5 & 9.82 $\pm$ .4 & 11.05 $\pm$ .38 \\ 
ds-reasoner & 8 & 24 & 8.0 $\pm$ .42 & 6.83 $\pm$ .16 & 59.99 $\pm$ 8.28 & 58.19 $\pm$ 4.83 \\
o4-mini & 6 & 18 & 7.5 $\pm$ .34 & 4.33 $\pm$ .16 & 23.88 $\pm$ 5.26 & 24.62 $\pm$ 3.48 \\
ds-chat & 1 & 11 & 12.0 $\pm$ .0 & 10.27 $\pm$ .65 & 4.91 $\pm$ .0 & 6.61 $\pm$ .24 \\

\bottomrule
\end{tabular}
\vspace{-1em}
\end{table}

\begin{figure}[ht]
\centering
\includegraphics[width=\linewidth]{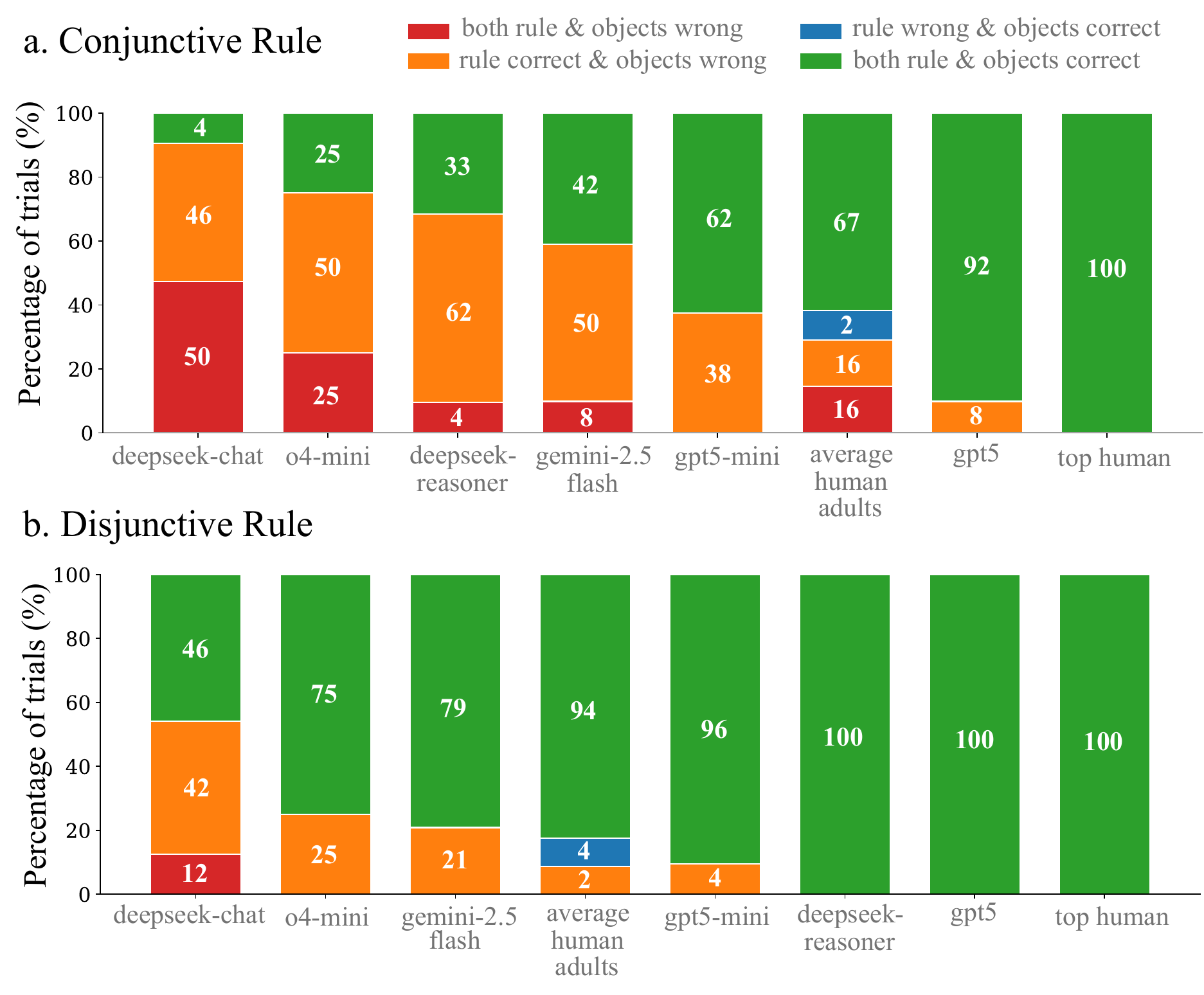}
\vspace{-7mm}
\caption{\textbf{Distribution of rule and object identification outcomes} for conjunctive and disjunctive conditions in adults and LLMs with active exploration. \vspace{-5mm}}
\label{fig:dist_of_mistakes}
\end{figure}

\vspace{-4mm}
\paragraph{Conjunctive vs.\ Disjunctive Reasoning.}
Our finding shows that adults perform a greater number of tests and actions in conjunctive case  compared to disjunctive case, and this pattern is similarly observed  across language models (see Table~\ref{tab:exploration-all}). 
Table~\ref{tab:exploration-success} further shows that even successful participants and models required more tests to converge on the correct causal hypothesis in conjunctive settings. These findings suggest that the challenge of conjunctive reasoning lies not simply in representing conjunctive relations, but in resolving ambiguity among competing hypotheses.
In a disjunctive world ($A\lor B$), any single positive test, (e.g. testing $A$ and seeing it work) provides immediate, high certainty evidence. In contrast, in a conjunctive world ($A\land B$), a single successful test (testing $A$,$B$ together) is insufficient to rule out $A$ alone, $B$ alone, or $A\land B$. This suggests that conjunctive conditions require more data points to reach the same level of hypothesis pruning, see Figure~\ref{fig:hypothesis_infogain}. 
The steep drop in the  number of hypothesis remaining in Figure~\ref{fig:hypothesis_infogain} suggests that this exploration is not random. Like the children in \citep{schulz2007preschool, pmlr-v177-kosoy22a}, adults are performing rather efficient experiments to prune the hypothesis space. 
\vspace{-5mm}
\paragraph{Contingent Feedback vs. Intervention Planning.}
To isolate whether the benefits of active exploration arise from intervention planning itself or from receiving contingent feedback from one’s own interventions, we introduced a third participant group: Passive Proposers. Passive Proposers generated interventions in the same manner as active participants, but did not directly observe the outcomes of their proposed tests. Instead, each Passive Proposer received the outcome produced by a matched Active Explorer’s intervention sequence. Thus, Passive Proposers engaged in hypothesis generation and intervention planning without maintaining direct control over evidence generation. In conjunctive environments, Passive Proposers performed substantially worse on object identification accuracy ($0.12$) than both Passive Observers ($0.45$) and Active Explorers ($0.69$), despite actively reasoning about the hypothesis space. Rule selection accuracy showed a similar but weaker pattern (Passive Proposer: $0.57$, Passive Observer: $0.69$, Active Explorer: $0.82$). In disjunctive environments, performance was generally higher across all conditions, although Active Explorers still achieved the strongest performance overall. 
All reported numbers here are mean performance across participants.
These results suggest that the benefits of active exploration do not arise solely from internally generating hypotheses or planning interventions. Rather, successful causal discovery appears to depend critically on the tight coupling between intervention selection and contingent outcome observation.
\vspace{-5mm}
\paragraph{Error Structure Across Rule and Object Inference.} 
Figure~\ref{fig:dist_of_mistakes} decomposes performance into four jointly defined outcome categories, reflecting whether participants correctly inferred the causal rule and identified the causal object(s). 
Across both conjunctive and disjunctive conditions, the majority of human participants successfully inferred both the underlying rule and the correct causal objects. However, the structure of errors differs substantially between the two causal environments. In the disjunctive condition, errors were relatively rare and typically partial:  human participants usually identified either the correct rule or the correct objects, with few simultaneous failures on both dimensions. Most state-of-the-art models similarly showed strong disjunctive performance, with \texttt{gpt-5} and \texttt{deepseek-reasoner} achieving near-perfect or perfect full-hypothesis accuracy.
The conjunctive condition revealed a qualitatively different error profile. Although many participants still correctly inferred both the rule and the causal objects, failures were more common and heterogeneous. Among average human adults, 16\% incorrectly inferred both the rule and the objects, while another 16\% correctly inferred the conjunctive rule but misidentified the causal objects. This pattern suggests that participants often recognized the need for a conjunctive explanation while still struggling to uniquely determine the correct causal set. Interestingly, when language models made object-identification errors, these failures were most often driven by misclassifying only a single causal object, a near-miss pattern that was comparatively less common in human participants.

\vspace{-2mm}

\vspace{-3mm}
\paragraph{Complexity of Hypothesis Space and Search Strategy.}
The analysis of information gain patterns reveals systematic differences in how humans and language models resolve uncertainty across conjunctive and disjunctive environments. In the disjunctive condition, participants accumulate most available information within only a few tests, and the marginal utility of additional exploration rapidly approaches zero once the rule is identified (Figure~\ref{fig:hypothesis_infogain}).  In contrast, conjunctive environments require substantially longer exploration trajectories, with informative interventions continuing to emerge later in the sequence. This pattern is consistent with the behavioral observation that adults perform more tests in conjunctive settings ($M$=9.6) than in disjunctive settings ($M$=6.4; see Table~\ref{tab:exploration-all}), reflecting the greater ambiguity of conjunctive hypothesis spaces.
Figure~\ref{fig:search_strategy} further shows that humans and language models adapt their intervention strategies to the underlying causal structure. In disjunctive environments, both humans and stronger models rely primarily on sparse interventions involving one or two objects, which are often sufficient to identify both the rule and causal objects. In conjunctive environments, however, agents progressively transition toward multi-object interventions because single-object tests are insufficient to disambiguate competing explanations. Average human adults gradually shift from one-object to two-object interventions, while top-performing humans rapidly converge on targeted multi-object combinations. Stronger reasoning-oriented models such as \texttt{gpt-5} exhibit partially similar exploration trajectories, dynamically adjusting intervention complexity in response to remaining uncertainty.

\vspace{-2mm}

\vspace{-4mm}
\paragraph{Human and LLM Performance in Causal Discovery.}
Figure~\ref{fig:llm-human-bar-active} compares full hypothesis accuracy across humans and language models using our active exploration setup. Performance varies considerably across both models and causal structures. In the disjunctive condition, several state-of-the-art models achieve near-human performance. \texttt{gpt-5} and \texttt{deepseek-reasoner} perform as well as top-performing human participants. These results suggest that disjunctive causal environments are comparatively easy to resolve once agents are able to actively generate informative interventions.
The conjunctive condition reveals substantially larger performance differences across agents. Although top-performing humans continue to achieve perfect accuracy and average human adults substantially outperform most models, many language models exhibit pronounced declines in conjunctive reasoning performance. Smaller or less structured models such as \texttt{deepseek-chat} and \texttt{o4-mini} perform only modestly above baseline, whereas stronger reasoning-oriented models such as \texttt{gpt-5} and \texttt{gpt5-mini} show substantially greater robustness. Nevertheless, even strong models generally remain below top human performance in conjunctive settings.
 \vspace{-4mm}

\vspace{-5mm}
\section{Discussion}
The present findings suggest that adults’ apparent failures in conjunctive causal learning arise less from a fixed limitation in causal reasoning and more from the structure of the learning environment. Under passive observation, adults may over-rely on disjunctive priors, consistent with prior demonstrations of a conjunctive handicap \citep{lucas2014children}. Yet when allowed to actively explore, adults generated informative interventions and successfully inferred conjunctive rules and causal objects, demonstrating a flexibility more commonly associated with young children \citep{gopnik2015younger}.

\noindent Our reinterpretation aligns with interventionist accounts of causal learning, which treat causal knowledge as understanding how actions change the world \citep{cook2011science,woodward2003making,yiu2025empowerment}. Passive observation paradigms constrain learners to reason over experimenter-selected evidence, which limits their ability to design tests that directly adjudicate between competing hypotheses they may hold in their minds, leaving familiar disjunctive priors intact rather than prompting the learners to revise them. By contrast, active exploration lets learners test the hypotheses that they themselves are considering. In our task, adults used this to propose tests that narrow their own hypothesis space, generating informative interventions that supported successful inference of conjunctive rules where passive observations failed.

\noindent Importantly, the reported improvement in conjunctive causal learning was not cost-free. Although time per test was comparable across rule types, participants conducted \textit{more} tests in the conjunctive condition than in the disjunctive condition. This suggests that conjunctive discovery requires a more extended and strategic search process because conjunctive structures require ruling out multiple simpler alternative explanations \citep{sobel2004children}. Thus, active exploration does not eliminate the informational demands of conjunctive learning, but instead allows adults to reach those demands through targeted intervention rather than heuristic shortcuts or priors. Active exploration reveals adult competence not by making the task easier, but by giving learners the means to gather the evidence they need.

\noindent The follow-up Passive Proposer experiment further clarified the source of the active exploration advantage. Although Passive Proposers actively generated hypotheses and proposed interventions, they did not receive contingent feedback from their own actions. Instead, they observed outcomes generated by another participant’s intervention sequence, and consequently performed substantially worse than Active Explorers. This suggests that active learning benefits are not reducible to explicit hypothesis generation or test selection alone. They depend critically on learners being able to observe the outcome of the intervention they choose.

%



\noindent These results also help reconcile prior developmental results. Children’s apparent advantage in conjunctive causal learning has been attributed to weaker prior commitments or greater openness to unlikely hypotheses \citep{gopnik2015younger}. Our results suggest that adults \textit{do} possess the representational capacity to infer unlikely structures like conjunctive rules too, but they often rely on efficient default heuristics that perform well under sparse or passively presented evidence. When adults are placed in an environment that supports hypothesis-driven exploration, their causal learning resembles that of younger children in its flexibility and sensitivity to evidence. This pattern suggests that previously reported developmental differences  reflect differences in the strength of default heuristics and willingness to revise beliefs, and not fundamental differences in causal reasoning mechanisms. 

\noindent The present results also clarify how adult performance compares to large language models under active exploration. Our result shows that model performance is heterogeneous: some actively exploring models perform poorly in both disjunctive and conjunctive settings, while others like \texttt{gpt-5} achieve accuracy that is close to that of actively exploring adults across both rule types. 
Together, these differences suggest that while some models can match humans on full hypothesis accuracy, key aspects of strategic testing, evidence selection, and rule-level inference may still distinguish human causal learning.

\noindent Collectively, these findings demonstrate that adult causal learning is more adaptive and flexible than prior passive paradigms might have suggested. Apparent failures in conjunctive reasoning may arise not from an inability to represent complex causal structure, but from the interaction between strong default heuristics and constrained learning environments. Allowing adults to act like "scientists" by exploring and generating evidence reveals forms of causal competence that passive observation can obscure. By contrast, the LLM comparison suggests that active intervention alone is insufficient for human-like causal discovery: successful performance depends on the quality, efficiency, and informativeness of the tests, particularly in conjunctive environments.


\section{Acknowledgements}

This research was enabled in part by support provided by (Calcul Québec) (https://www.calculquebec.ca/en/) and the Digital Research Alliance of Canada (https://alliancecan.ca/en). We acknowledge support by NSERC (Discovery Grant: RGPIN-2020-05105; Discovery Accelerator Supplement: RGPAS-2020-00031) and CIFAR (Canada AI Chair; Learning in Machine and Brains Fellowship).

\printbibliography

\end{document}